\documentclass[runningheads]{llncs}

\usepackage[T1]{fontenc}
\usepackage{amsmath}
\usepackage{booktabs}
\usepackage{amssymb}
\usepackage{graphicx}
\usepackage{float}
\usepackage{placeins}
\usepackage[table]{xcolor}

\definecolor{OursRow}{HTML}{EDF4FA}

\begin{document}
\raggedbottom

\title{Measured Sliders: Learning Continuous Controls from Differentiable Image Measurements}
\titlerunning{Measured Sliders}

\author{Yijia Chen \and Boyu Wei \and Xuanhua Yin\thanks{Corresponding author.}}
\authorrunning{Yijia Chen et al.}

\institute{School of Computer Science, The University of Sydney\\
\email{yche0587@uni.sydney.edu.au, bwei0951@sydney.edu.au,
xuanhua.yin@sydney.edu.au}}

\maketitle

\begin{abstract}
Continuous sliders are useful only when coefficient changes produce
predictable image changes.  Yet most diffusion sliders derive their axes from
text or learned representations, leaving their scales disconnected from
observable image properties.  Consequently, we cannot tell in advance which
attributes are learnable, compare control strengths directly, or anticipate
interference when multiple controls are combined.  We propose
\emph{Measured Sliders}, a framework that defines continuous controls through
closed-form differentiable image measurements.  A common measurement space
unifies the pipeline.  Before training, an observability test identifies usable
supervision.  During training, a measurement-guided objective learns target
movement while suppressing non-target changes.  After training, decoded
calibration expresses controls in comparable units of realized image change.
Multiple LoRA branches are stored
in one checkpoint and composed without training on joint activations.  Across
SDXL and FLUX.1-dev, the resulting controls are ordered, selective, and
composable.  On 553 prompts, lighting direction reaches $\rho=0.995$ and
$98.9\%$ monotone sweeps.  A five-attribute checkpoint achieves average
selectivity $2.59$, compared with $1.50$ for the strongest baseline, and
preserves every requested direction in $96.7\%$ of pair and $86.1\%$ of triple
compositions.  The observability test also separates every subsequently
successful measurement from the failed candidate.  Overall, image-space
measurement provides a common basis for learning, diagnosing, calibrating, and
composing continuous generative controls.
\end{abstract}

\keywords{Diffusion models \and Controllable generation \and
          Attribute control \and Evaluation protocols}

\begin{figure}[h!]
  \centering
  \includegraphics[width=\textwidth]{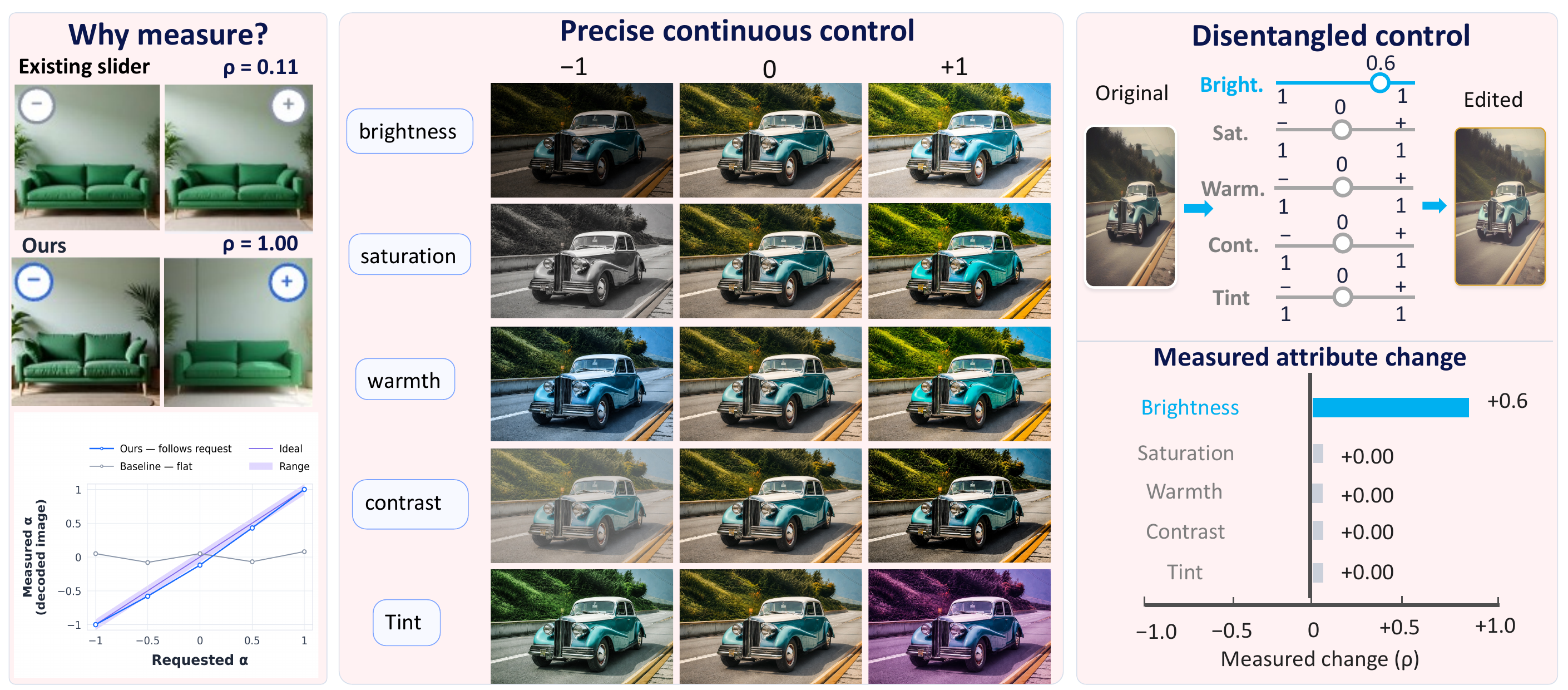}
  \caption{\textbf{Measured Sliders turn image measurements into precise,
  disentangled controls.} Left: in the matched FLUX setting, the text-defined
  slider is nearly flat, whereas measurement supervision follows the requested
  direction. Center: one checkpoint provides continuous brightness, saturation,
  warmth, contrast, and tint axes. Right: activating brightness produces the
  requested decoded change while the other four measurements remain nearly
  unchanged.}
  \label{fig:teaser}
\end{figure}

\section{Introduction}
\label{sec:intro}

Text-to-image diffusion models generate high-quality content, yet fine-grained
creation also requires continuous control of properties such as brightness,
saturation, color temperature, and lighting direction while preserving subject,
geometry, and composition~\cite{rombach2022high,sdxl,fang2024camera,yuan2025generative,chen2025attrictrl}.
Text can name an edit direction such as ``slightly brighter,'' but it does
not specify how much change should appear in the generated
image~\cite{gandikota2024sliders,baumann2025attribute,chiu2026textslider,ezra2025freesliders}.
The distinction is especially important for scene-level attributes, whose
control requires coordinated changes across the image without changing the
scene itself~\cite{zeng2024dilightnet,zhang2024lumisculpt,lu2025intrinsic}.

Reusable sliders commonly learn a direction in parameter, embedding, or score
space from opposing prompts, semantic representations, or editing instructions,
and expose a raw coefficient $\alpha$ at inference
time~\cite{gandikota2024sliders,baumann2025attribute,chiu2026textslider,ezra2025freesliders,zarei2026slideredit}.
Unified variants store or compose several semantic controls in one
model~\cite{ye2026all,zarei2026slideredit}, numerical-conditioning methods encode
camera settings or intensities as tokens or embeddings~\cite{fang2024camera,yuan2025generative,chen2025attrictrl},
and output-aware guidance optimizes individual sampling
trajectories~\cite{dhariwal2021guidance}.  These approaches broaden how controls
are represented and combined, while the relation between their native
coordinate and the realized image change remains implicit.  Consequently,
identical raw coefficients across attributes or methods need not represent
comparable effects.

This gap affects learning and evaluation in three ways.  First, increasing a
coefficient need not yield an ordered or predictable response.  Range,
smoothness, and alignment can only diagnose the trajectory after it has been
learned~\cite{gandikota2024sliders,baumann2025attribute,ezra2025freesliders,zarei2026slideredit}.
Second, compositional objectives reduce semantic interference but do not
directly specify a measured target change together with measured non-target
invariance~\cite{ye2026all,zarei2026slideredit}.  Third, LoRA scales, embedding
offsets, and guidance weights have no shared output unit, so equal-coefficient
comparisons confound control quality with effect magnitude~\cite{baumann2025attribute,ezra2025freesliders}.
When training uses an inexpensive differentiable preview, an additional
question arises: whether that preview retains enough information about a
candidate decoded-image measurement to supervise it reliably.

Image-level guidance demonstrates that differentiable output objectives can
steer diffusion~\cite{dhariwal2021guidance}, while multi-attribute sliders make
interference and preservation explicit concerns~\cite{ye2026all,zarei2026slideredit,ezra2025freesliders}.
Together they motivate defining a reusable control in the same measurement
space in which its effect is evaluated.  A vector of image measurements can
state the desired displacement of one attribute and the invariance of the
others in a single objective.  Normalizing those measurements by natural image
variation further provides a common effect scale across heterogeneous
attributes.

We therefore propose \textbf{Measured Sliders}, a framework that converts
differentiable image measurements into reusable continuous controls.  As shown
in Fig.~\ref{fig:teaser}, each LoRA branch learns a target displacement in one
measurement while suppressing changes in the remaining measurements, and all
branches are stored in one checkpoint.  Before optimization, a pre-training
observability analysis compares preview error with decoded natural variation to
determine whether the training path preserves a usable signal.  After optimization,
decoded calibration maps raw coefficients to comparable units of realized
image change.  The same measurement space thus connects feasibility, learning,
and evaluation.

Experiments on SDXL and FLUX.1-dev provide four complementary findings.  On 553
prompts, Measured Sliders reaches $\rho=0.995$ for lighting direction, $98.9\%$
monotone sweeps, and $3.3\times$ the selectivity of the strongest external
slider.  In a controlled FLUX comparison that changes only the supervision,
measurement supervision increases realized effect by $295\times$ and raises
monotone sweeps from $9.6\%$ to $99.8\%$.  A jointly trained five-attribute
checkpoint attains average selectivity $2.59$ versus $1.50$ for the strongest
baseline average.  Simultaneous activation moves every requested attribute in
the intended direction in $96.7\%$ of pairs and $86.1\%$ of triples.  Finally,
pre-training observability predicts all six evaluated learnability outcomes, and
decoded calibration reduces endpoint dispersion from $63.2\times$ to
$2.2\times$. Our contributions are summariezed as follows:
\begin{itemize}
  \item We define a continuous slider by decoded-image measurements and jointly
  learn multiple LoRA branches with an objective that specifies target effect
  and non-target stillness.
  \item We use pre-training observability to identify usable measurements,
  natural-variation scaling to balance their optimization, and decoded
  calibration to express the resulting controls in comparable effect units.
  \item We evaluate the learned controls on SDXL and FLUX.1-dev in terms of
  decoded effect, drift, ordering, consistency, and zero-shot composition.
  Measured Sliders reaches $\rho=0.995$ with $98.9\%$ monotone lighting sweeps
  and preserves every requested direction in $96.7\%$ of pair compositions.
\end{itemize}

\section{Related Work}
\label{sec:related}

\subsection{Continuous and Composable Control}

Continuous diffusion control is commonly expressed as traversal in parameter
or representation space.  Concept Sliders learns low-rank parameter directions
from opposing prompts or paired images~\cite{gandikota2024sliders}.
Subject-specific control identifies directions in CLIP text
representations~\cite{baumann2025attribute}.  SliderSpace discovers multiple
interpretable LoRA directions from the distribution induced by one
prompt~\cite{gandikota2025sliderspace}.  SliderEdit decomposes compound editing
instructions into independently weighted components~\cite{zarei2026slideredit},
while PairEdit learns hard-to-verbalize transformations from paired source and target
examples~\cite{lu2025pairedit}.  These methods establish continuous semantic
coordinates across several model spaces.

Other editors obtain graded behavior without a persistent scalar parameter
direction.  Attribute Diffusion samples diverse attribute variations and
supports controlled exploration~\cite{parihar2025attribute}.  Stable Flow edits
through selected transformer features~\cite{avrahami2025stable}.  FeedEdit
regulates guidance from the evolving edit degree~\cite{fu2025feededit}.
UniEdit-I adds semantic verification~\cite{bai2026uniediti}, while
visual-example methods infer edits from paired images rather than
language~\cite{lu2025pairedit,elezabi2026languagefree}.

Recent work also increases the number and composability of controls.  All-in-One
Slider stores sparse text-embedding directions in one module~\cite{ye2026all},
and CompSlider learns conditional priors for simultaneous multi-attribute
generation~\cite{zhu2025compslider}.  Unified editors and adapters cover many
generation and editing tasks~\cite{fluxkontext2025,xia2025dreamomni,yu2025anyedit}, while
Conditional Balance selects condition-sensitive layers to trade off competing
inputs~\cite{cohen2025conditional}.  Prior work improves preservation with
causal representations~\cite{huang2025causal}, separate subject and context
control~\cite{wang2025psdiffusion}, and text conditioning designed to prevent
leakage~\cite{mun2025leakage}.  Measured Sliders defines its coordinate by
decoded-image measurements, placing target displacement and named non-target
invariances in the same space.

\subsection{Measurement-Aware Learning and Evaluation}

Differentiable output objectives provide another route to controllable
generation.  Classifier guidance modifies denoising with auxiliary
gradients~\cite{dhariwal2021guidance}.  Universal Guidance accepts general
differentiable guidance functions~\cite{bansal2023universal}.  FreeDoM uses
time-independent energies from pretrained networks~\cite{yu2023freedom}.
ADMMDiff decouples the diffusion prior and differentiable guidance
loss~\cite{zhang2025admm}.  Constrained diffusion learning combines multiple
reward and distributional constraints through joint primal and dual
optimization~\cite{khalafi2025constrained}.  Direct clean-image supervision
accelerates spatially controllable generation~\cite{sangare2026x0}.
These approaches demonstrate that image-level measurements can guide sampling
or distribution learning.  Our setting uses the measurement to construct a
persistent LoRA coordinate, which makes the same output quantity available for
training, deployment, and evaluation.  Before training, we additionally test
whether the inexpensive preview used for gradients preserves the decoded
variation of each candidate measurement.  This candidate-specific observability analysis
is distinct from factor-level identifiability results for weakly supervised
diffusion models~\cite{wang2025disentangle}.

Continuous-control evaluation measures response, monotonicity, preservation,
and interference~\cite{baumann2025attribute,zhu2025compslider}.
FreeSliders further detects saturation and reparameterizes nonlinear traversal
within one axis~\cite{ezra2025freesliders}.  General image-editing benchmarks
evaluate perceptual quality, instruction alignment, preservation, and semantic
stability~\cite{xu2025lmm4edit,wang2026i2ibench,li2026stablei2i}.  These criteria
assess whether an edit is effective and visually faithful.  We address the
orthogonal cross-axis question: coefficients from different controls are first
mapped to decoded natural-variation units, after which target response and
non-target drift can be compared at matched realized effect.

\section{Method}
\label{sec:method}

\begin{figure}[h!]
  \centering
  \includegraphics[width=\linewidth]{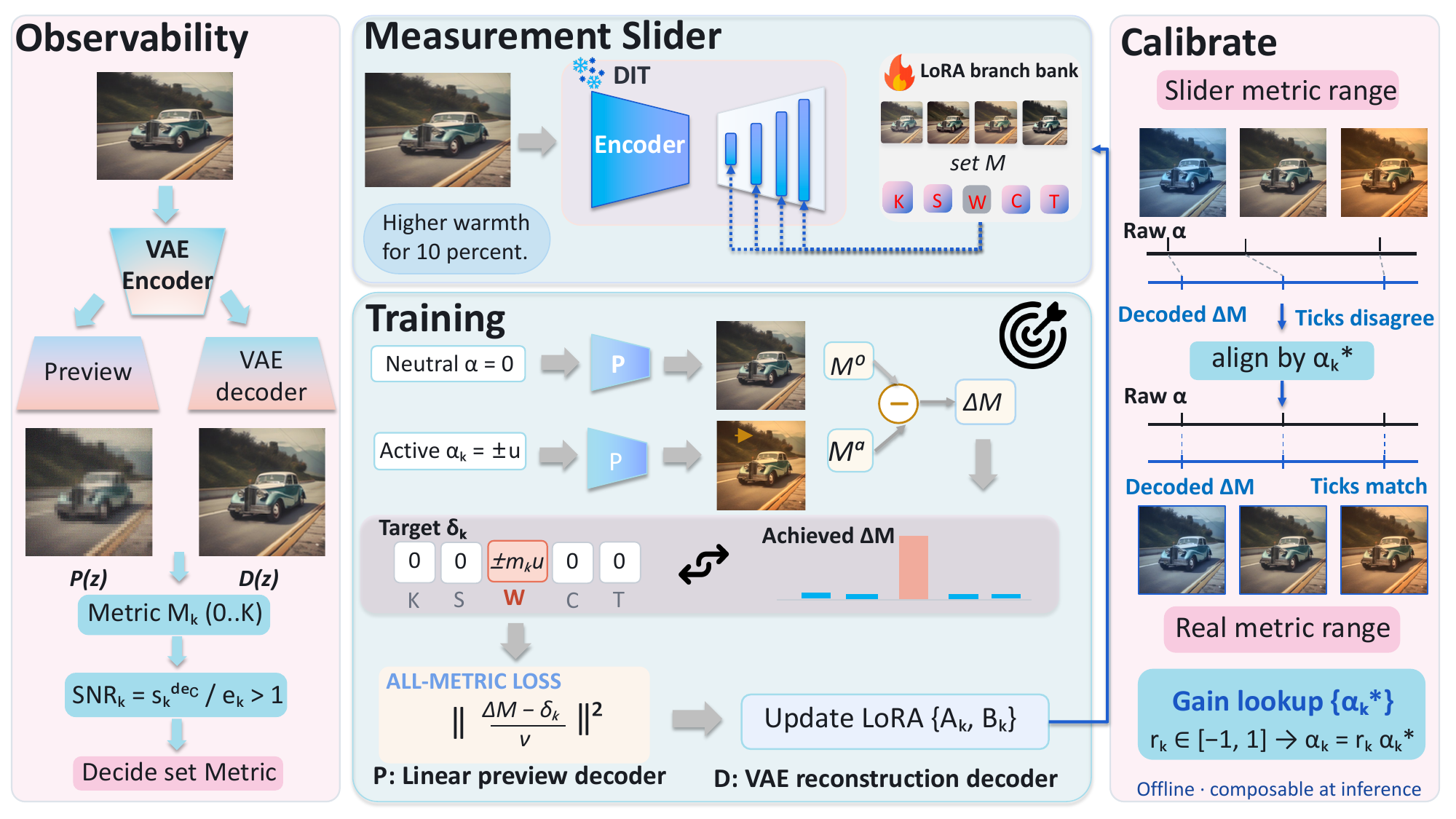}
  \caption{\textbf{Measured Sliders pipeline.} \emph{Observability:} preview and
  decoded measurements are compared before training to retain candidates with a
  reliable optimization signal. \emph{Train:} an active LoRA branch matches its
  target displacement while the remaining measurements are held still.
  \emph{Calibrate:} decoded responses map raw coefficients to comparable
  image-space effect units for inference.}
  \label{fig:method-pipeline}
\end{figure}

Measured Sliders constructs a control coordinate in three stages.  First,
\emph{pre-training observability analysis} determines whether the differentiable
preview used for optimization can resolve the natural variation of a candidate
measurement on decoded images.  Second, \emph{multi-control learning} trains branch-specific LoRA
updates against a shared vector of target and non-target measurements.  Third,
\emph{decoded calibration} converts each branch's raw coefficient into a common
image-space effect scale.  Figure~\ref{fig:method-pipeline} summarizes the
pipeline.

\subsection{Measurements and Pre-Training Observability}
\label{sec:preview}

\paragraph{Image measurements.}
For an RGB image $I$, we represent a panel of attributes by
\begin{equation}
\mathbf M^{q}(I)=
[M_1^{q}(I),\ldots,M_N^{q}(I)]^\top,
\qquad q\in\{\mathrm{pre},\mathrm{dec}\},
\label{eq:measurement-vector}
\end{equation}
\nopagebreak[4]
where $N$ is the number of controls and $q$ identifies the differentiable
preview or final decoded implementation.  We use closed-form, parameter-free
statistics, where $K$, $S$, $W$, $C$, $T$, and $L$ denote brightness,
saturation, warmth, contrast, tint, and spatial lighting direction.  In particular,
$M_L(I)=\langle Y\rangle_{\rm right}-\langle Y\rangle_{\rm left}$, where
$Y=0.299I_R+0.587I_G+0.114I_B$ is luma and $\langle\cdot\rangle$ denotes a spatial
mean.  The remaining measurements are
$M_S=\langle(\max_bI_b-\min_bI_b)/\max(\max_bI_b,\epsilon_S)\rangle$,
$M_W=\langle I_R-I_B\rangle$, $M_C=\operatorname{std}(Y)$, and
$M_T=\langle(I_R+I_B)/2-I_G\rangle$, where $b\in\{R,G,B\}$ and $\epsilon_S$
prevents division by zero.  Brightness combines mean luma with the fraction of
pixels above the fixed shadow threshold $\tau_d=0.2$.  Its preview replaces the
hard threshold by $\operatorname{sigmoid}(20(\tau_d-Y))$.

\paragraph{Differentiable preview.}
Decoding a clean-latent estimate at every training step is expensive.  We
therefore fit and freeze the affine preview
\begin{equation}
P(z)=\operatorname{clip}_{[0,1]}(W_P^\top z+b_P),
\label{eq:preview}
\end{equation}
\nopagebreak[4]
where $z\in\mathbb R^{C_z\times H\times W}$ is an unpacked latent,
$W_P\in\mathbb R^{C_z\times3}$ and $b_P\in\mathbb R^3$ are obtained by ridge
regression against downsampled decoded images, and $P(z)$ is an RGB preview at
latent resolution.  Training uses $\mathbf M^{\rm pre}(P(z))$.  Observability analysis,
calibration, and evaluation use $\mathbf M^{\rm dec}(D(z))$, where $D$ is the
frozen VAE decoder.

\paragraph{Pre-training observability.}
A preview is useful only when its error is smaller than the decoded variation
that it must resolve.  For candidate $k$, we compute
\begin{equation}
\begin{aligned}
\varepsilon_k^2
&=\frac{1}{|\mathcal H|}\sum_{z\in\mathcal H}
  \big[M_k^{\rm pre}(P(z))-M_k^{\rm dec}(D(z))\big]^2,\\
s_k
&=Q_{95}\!\left(\mathcal V_k\right)-Q_{5}\!\left(\mathcal V_k\right),
\quad
\mathrm{SNR}_k=\frac{s_k}{\varepsilon_k},
\end{aligned}
\label{eq:observability}
\end{equation}
\nopagebreak[4]
where $\mathcal H$ is a held-out latent set,
$\mathcal V_k=\{M_k^{\rm dec}(D(z)):z\in\mathcal S_{\rm ref}\}$ contains
measurements from an independent neutral reference set, $s_k$ is its robust
decoded range, $Q_{95}$ and $Q_5$ denote the indicated percentiles, and $\mathrm{SNR}_k$ is the
pre-training observability ratio.  We
retain a candidate when $\mathrm{SNR}_k>1$, meaning that decoded natural variation exceeds preview
RMSE.  This test uses only the frozen preview, decoder, and reference images.

\subsection{Measurement-Supervised Multi-Control Learning}
\label{sec:objective}

\paragraph{Multi-branch controller.}
Let $f_\theta^{\boldsymbol\alpha}(z_t,t,c)$ be a frozen diffusion backbone with
control vector $\boldsymbol\alpha=(\alpha_1,\ldots,\alpha_N)$, where $z_t$ is a
noisy latent, $t$ is the diffusion timestep, and $c$ is the condition.  Each
selected linear projection is augmented as
\begin{equation}
W(\boldsymbol\alpha)x
=W_0x+\gamma\sum_{k=1}^{N}\alpha_kB_kA_kx,
\label{eq:slider}
\end{equation}
\nopagebreak[4]
where $x$ is the input activation, $W_0$ is frozen,
$A_k\in\mathbb R^{r_{\rm L}\times d_{\rm in}}$ and
$B_k\in\mathbb R^{d_{\rm out}\times r_{\rm L}}$ are the rank-$r_{\rm L}$ LoRA factors of branch
$k$, and $\gamma$ is the fixed LoRA scale.  We zero-initialize $B_k$, so
$\boldsymbol\alpha=\mathbf0$ exactly recovers the backbone.  All branches are
optimized in one run and stored in one checkpoint, while their factors remain
branch-specific.

Let $\mathcal R_t$ recover a clean-latent estimate from the backbone output and
define
\begin{equation}
\begin{aligned}
\widehat z_0^{\boldsymbol\alpha}
&=\mathcal R_t\!\left(z_t,
f_\theta^{\boldsymbol\alpha}(z_t,t,c_0)\right),\\
\Delta\mathbf M(\boldsymbol\alpha)
&=\mathbf M^{\rm pre}\!\left(P(\widehat z_0^{\boldsymbol\alpha})\right)
-\mathbf M^{\rm pre}\!\left(P(\widehat z_0^{\mathbf0})\right),
\end{aligned}
\label{eq:deltaM}
\end{equation}
\nopagebreak[4]
where $c_0$ is a neutral content condition and
$\Delta\mathbf M\in\mathbb R^N$ is the preview measurement displacement from
the frozen-model output.

\paragraph{Effect-and-stillness objective.}
At each update, we sample branch $k$, polarity $\pi\in\{-1,+1\}$, and magnitude
$u\sim\mathcal U(u_{\min},u_{\max})$, then set
$\boldsymbol\alpha=\pi u\mathbf e_k$.  The measurement loss is
\begin{equation}
\mathcal L_{\rm meas}
=\left\|\operatorname{diag}(\boldsymbol\nu)^{-1}
\left[\Delta\mathbf M(\pi u\mathbf e_k)
-\pi u m_k\mathbf e_k\right]\right\|_2^2,
\label{eq:lmetric}
\end{equation}
\nopagebreak[4]
where $\mathbf e_k$ is the $k$th standard basis vector, $m_k>0$ is the requested
movement of measurement $k$, and
$\boldsymbol\nu=(\nu_1,\ldots,\nu_N)$ contains fixed training normalizers.
The $k$th residual specifies the target effect.  Every other residual has a zero
target and therefore specifies non-target stillness.  Normalization prevents a
large-unit measurement from dominating the joint objective.

The complete loss is
\begin{equation}
\mathcal L=\lambda_{\rm m}\mathcal L_{\rm meas}
+\lambda_{\rm a}\mathcal L_{\rm anchor}
+\lambda_{\rm cb}\mathbf1[k=K]\mathcal L_{\rm cb},
\label{eq:total}
\end{equation}
\nopagebreak[4]
where $\mathcal L_{\rm anchor}$ is an optional frozen-text consistency term,
$\mathcal L_{\rm cb}$ preserves color balance for brightness, and the
$\lambda$ terms are their weights.  These terms select a content-preserving
realization without redefining the measurement target.  Specifically, the
anchor matches the controlled prediction to the neutral prediction plus a
frozen direction between the positive and negative text conditions.  The brightness term penalizes the
$\ell_1$ difference between controlled and neutral mean chroma
$\langle I-Y(I)\mathbf1\rangle_{h,w}$.

\subsection{Decoded Calibration and Multi-Control Interface}
\label{sec:calib}

Raw LoRA coefficients are specific to a branch.  We evaluate a positive ladder
$\mathcal A=\{0<a_1<\cdots<a_J\}$ on a fixed calibration set $\mathcal C$ and
compute
\begin{equation}
\overline d_k(a)=\frac{1}{|\mathcal C|}\sum_{\xi\in\mathcal C}
\left[M_k^{\rm dec}(G_{a\mathbf e_k}(\xi))
-M_k^{\rm dec}(G_{\mathbf0}(\xi))\right],
\label{eq:ladder}
\end{equation}
\nopagebreak[4]
where $\xi$ pairs a prompt with a seed, $G_{\boldsymbol\alpha}(\xi)$ is its fully
decoded generation, and $\overline d_k(a)$ is the mean decoded response of
branch $k$.  Let $\widetilde d_k(a_j)=\max_{\ell\le j}\overline d_k(a_\ell)$
be the running-max response.  The calibrated endpoint is
\begin{equation}
\alpha_k^\star=\widetilde d_k^{-1}(\tau s_k),
\label{eq:calib}
\end{equation}
\nopagebreak[4]
where $\tau s_k$ is the desired decoded effect, $\tau>0$ is shared across
attributes, and the inverse denotes the earliest crossing of the linearly
interpolated running-max curve, clipped to $[a_1,a_J]$.  This construction avoids
extrapolation and prevents a local response reversal from selecting a later
crossing.

Finally, a user specifies $\mathbf r\in[-1,1]^N$ through
\begin{equation}
\boldsymbol\alpha(\mathbf r)
=(r_1\alpha_1^\star,\ldots,r_N\alpha_N^\star),
\label{eq:user-control}
\end{equation}
\nopagebreak[4]
where $r_k$ is the normalized position of slider $k$.  Calibration changes no
model weight and adds no denoising step.  One nonzero coordinate activates an
individual control.  Several nonzero coordinates compose the corresponding
branches without retraining.

\newcommand{\MainOursSelectivity}{1.761}
\newcommand{\MainOursCILow}{1.692}
\newcommand{\MainOursCIHigh}{1.832}
\newcommand{\MainBestSliderSelectivity}{0.537}
\newcommand{\MainImprovement}{3.3}
\newcommand{\CalibrationBeforeSpread}{63.2}
\newcommand{\CalibrationAfterSpread}{2.2}
\newcommand{\FluxResponseGain}{295}
\newcommand{\FluxSelectivityGain}{1.12}
\newcommand{\StillnessSelectivityGain}{2.38}
\newcommand{\AttributeOursAverage}{2.59}
\newcommand{\AttributeBestBaselineAverage}{1.50}
\newcommand{\AttributePanelGain}{1.73}
\newcommand{\AttributeMinEffect}{0.57}
\newcommand{\AttributeMinRho}{0.967}
\newcommand{\AttributeOursEffect}{0.92}
\newcommand{\AttributeBestBaselineEffect}{0.61}
\newcommand{\AttributeOursDrift}{0.42}
\newcommand{\AttributeBestEffectBaselineDrift}{0.42}
\newcommand{\AttributeOursDino}{0.944}
\newcommand{\AttributeBestBaselineDino}{0.930}
\newcommand{\ObservabilityAccuracy}{6/6}
\newcommand{\ObservabilityMinSNR}{2.82}
\newcommand{\ObservabilityMaxSNR}{9.13}
\newcommand{\ObservabilityMinRho}{0.967}
\newcommand{\ObservabilityNegativeSNR}{0.419}
\newcommand{\ObservabilityNegativeRho}{-0.232}

\begin{table}[!t]
\centering
\caption{\textbf{Lighting-direction control on 553 GenEval prompts.} All
statistics use final decoded images. Effect and drift use fixed
backbone-specific natural ranges and are not compared across panels.
Monotonicity is reported as a percentage.}
\label{tab:main}
\small
\setlength{\tabcolsep}{1.6pt}
\renewcommand{\arraystretch}{1.08}
\begin{tabular}{@{}lrrrrrr@{}}
\toprule
Method & Effect & Drift$\downarrow$ & Sel.$\uparrow$ & $\rho\uparrow$ &
Mono.$\uparrow$ & DINO$\uparrow$ \\
\midrule
\multicolumn{7}{@{}l}{\emph{SDXL: decoded natural-variation units}} \\
\midrule
Concept Sliders & 0.039 & 0.085 & 0.453 & 0.122 & 33.5 & 0.924 \\
Attr. Control & 0.106 & 0.214 & 0.496 & $-0.112$ & 18.4 & 0.810 \\
Attr. Control, learned & 0.101 & 0.254 & 0.395 & 0.117 & 24.6 & 0.818 \\
FreeSliders & 0.026 & 0.053 & 0.497 & $-0.035$ & 26.2 & 0.944 \\
Text Slider & 0.020 & 0.037 & 0.537 & $-0.077$ & 21.3 & \textbf{0.957} \\
\addlinespace[1pt]
Prompt ladder & 0.204 & 0.363 & 0.560 & N/A & N/A & N/A \\
\cellcolor{OursRow}\textbf{Measured Sliders (ours)} & \cellcolor{OursRow}0.153 & \cellcolor{OursRow}0.087 & \cellcolor{OursRow}\textbf{1.761} & \cellcolor{OursRow}\textbf{0.995} & \cellcolor{OursRow}\textbf{98.9} & \cellcolor{OursRow}\textbf{0.957}
\\
\midrule
\multicolumn{7}{@{}l}{\emph{FLUX.1-dev: fixed SDXL-derived decoded-reference units}} \\
\midrule
Text anchor & 0.010 & 0.015 & 0.716 & 0.111 & 9.6 & 0.992 \\
Attr. Control & 0.058 & 0.097 & 0.598 & $-0.037$ & 20.3 & 0.944 \\
Prompt ladder & 1.122 & 1.534 & 0.731 & 0.824 & 64.4 & 0.722 \\
FreeSliders & 1.949 & 0.652 & 2.990 & 0.984 & 98.2 & 0.785 \\
\cellcolor{OursRow}\textbf{Ours} & \cellcolor{OursRow}\textbf{3.070} & \cellcolor{OursRow}0.915 & \cellcolor{OursRow}\textbf{3.356} & \cellcolor{OursRow}\textbf{0.998} & \cellcolor{OursRow}\textbf{99.8} & \cellcolor{OursRow}0.792
\\
\bottomrule
\end{tabular}
\end{table}

\section{Experiments}
\label{sec:experiments}

\subsection{Experimental Setup}
\label{sec:setup}

\paragraph{Models and training.}
We use frozen SDXL~\cite{sdxl} and FLUX.1-dev~\cite{flux} backbones and train
rank-4 LoRA branches with AdamW at learning rate $10^{-4}$ on one NVIDIA RTX
5090 with 32\,GB memory.  The SDXL appearance checkpoint jointly stores $K$ for
brightness, $S$ for saturation, $W$ for warmth, $C$ for contrast, and $T$ for
tint.  It uses 270 $768^2$ images and 600 updates.  A separate SDXL branch uses
$L$ for lighting direction.  The FLUX checkpoint stores $L,K,S,C$ and uses 64 $512^2$
images and 800 cyclic updates.  The preview is fitted once from 60 images and
then frozen.  Appearance endpoints use four calibration contents and target
$\tau=0.6$ in Eq.~\eqref{eq:calib}.  SDXL augments 368 decoder-half attention
projections and uses neutralized captions with
$(\lambda_{\rm m},\lambda_{\rm a},\lambda_{\rm cb})=(4,1,6)$.  FLUX augments 97
later-half attention projections, uses empty conditioning, and optimizes the
measurement loss alone.

\paragraph{Benchmarks and baselines.}
Lighting direction is evaluated on all 553 GenEval prompts~\cite{geneval} with
one seed and five coefficient levels.  The appearance benchmark uses 12
contents, two seeds, and seven levels: six contents select baseline operating
ranges and the disjoint six form the test split.  We compare official Concept
Sliders~\cite{gandikota2024sliders}, AttributeControl~\cite{baumann2025attribute},
Text Slider~\cite{chiu2026textslider}, FreeSliders~\cite{ezra2025freesliders},
and an ordered prompt ladder.  Baseline ranges maximize selectivity on the
selection split subject to mean DINO similarity $\geq0.80$, then remain fixed on
the test split.  All methods use two seeds except the archived FreeSliders run,
which uses one.  Zero-shot composition enumerates all ten pairs and ten triples,
all sign patterns, and both seeds without composition training.
In the FLUX supervision comparison, the text-anchor control matches our
architecture, data, optimizer, and update budget.  Only the training objective
changes.

\paragraph{Metrics and statistics.}
For driven attribute $k$, effect is $e_k=\Delta_k/s_k$, where $\Delta_k$ is its
decoded sweep range and $s_k$ is the decoded natural range from
Eq.~\eqref{eq:observability}.  Drift is
$d_k=\max_{j\ne k}\Delta_j/s_j$, where the maximum spans the four non-target
appearance measurements for appearance control and all five for lighting.
Selectivity is $R=\mathbb E[e_k]/\mathbb E[d_k]$.  We also report Spearman
correlation $\rho$, the percentage of monotone sweeps with $\rho>0.8$, and mean
pairwise DINO cosine similarity~\cite{dino}.  Statistics are aggregated per
combination of prompt and seed.  The 95\% confidence intervals use 10,000 bootstrap resamples and
cluster two-seed experiments by content.  Effect units are fixed within each
result panel and are not compared across backbones.  For composition, all-sign
success requires every active measurement to move in its requested direction.
we additionally report the weakest active effect, additivity error, inactive
drift, and DINO similarity.

\subsection{Main Results}
\label{sec:main}

\paragraph{Measurement supervision produces an ordered, selective axis.}
In the matched FLUX comparison, replacing the text prediction-difference target
with Eq.~\eqref{eq:lmetric} increases decoded lighting response by
$\FluxResponseGain\times$, raises $\rho$ from $0.111$ to $0.998$, and increases
monotone sweeps from $9.6\%$ to $99.8\%$, as reported in Table~\ref{tab:main}.  On SDXL,
Measured Sliders reaches selectivity $\MainOursSelectivity$, a
$\MainImprovement\times$ gain over the strongest external slider, with
$\rho=0.995$ and $98.9\%$ monotone sweeps.  Its selectivity has 95\% CI
$[\MainOursCILow,\MainOursCIHigh]$.  The prompt ladder has greater raw
movement but $4.2\times$ more non-target drift, with values of $0.363$ and
$0.087$, respectively.  The
result holds across a latent-diffusion UNet and a flow-matching transformer.

\begin{figure}[t]
\centering
\includegraphics[width=\linewidth]{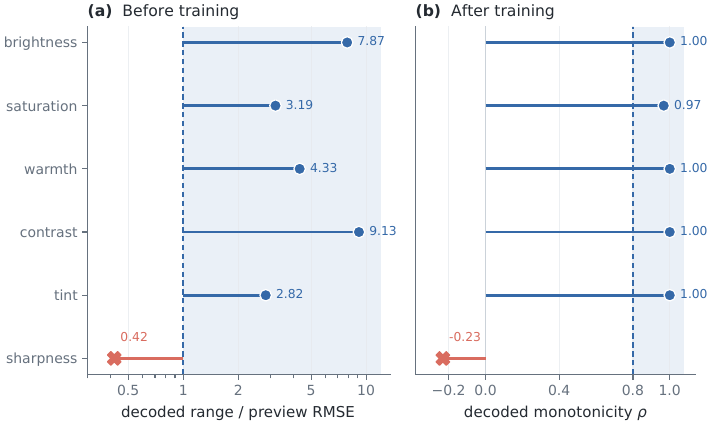}
\caption{\textbf{Pre-training observability predicts usable supervision.}
Each row pairs the observability ratio in the left panel with the subsequently
measured decoded monotonicity in the right panel.  Dashed lines mark the
fixed retention threshold $\mathrm{SNR}_k=1$ and learned-control threshold
$\rho=0.8$.  Five retained measurements exceed both thresholds.  Sharpness falls
below both.}
\label{fig:observability}
\end{figure}

\paragraph{Observability predicts usable supervision and calibration aligns effects.}
The five retained measurements have observability ratios
$\mathrm{SNR}_k\in[\ObservabilityMinSNR,\ObservabilityMaxSNR]$ and yield mean decoded
$\rho\geq\ObservabilityMinRho$ after training.  Sharpness has
$\mathrm{SNR}_k=0.419$ and produces an oppositely ordered response with
$\rho=-0.232$.  This outcome was predicted before its adapter was trained, as
shown in Fig.~\ref{fig:observability}.  Applying one preview-scaled coefficient to all
branches produces $\CalibrationBeforeSpread\times$ endpoint dispersion, whereas
decoded calibration on four separate contents reduces it to
$\CalibrationAfterSpread\times$ on the unseen split without changing model
weights.

\begin{table}[!b]
\centering
\caption{\textbf{Multi-control performance on six unseen contents.}
Upper: normalized selectivity of five branches in one checkpoint.  Lower:
zero-shot activation of all ten pairs and triples across every sign pattern and
two seeds.  All-sign success is a percentage.}
\label{tab:multicontrol}
\small
\setlength{\tabcolsep}{1.2pt}
\renewcommand{\arraystretch}{1.08}
\begin{tabular}{@{}lcccccc@{}}
\toprule
\multicolumn{7}{@{}l}{\textbf{Individually activated controls}}\\
Method & Bright. $K$ & Satur. $S$ & Warmth $W$ & Contrast $C$ & Tint $T$ & Avg. \\
\midrule
Concept Sliders & 0.47 & 1.19 & 0.81 & 0.76 & 0.55 & 0.76 \\
Attr. Control & 2.18 & 1.10 & 1.08 & 0.73 & 0.76 & 1.17 \\
FreeSliders$^{\dagger}$ & 1.41 & \textbf{1.47} & 1.61 & 0.86 & \textbf{2.14} & 1.50 \\
Text Slider & 1.81 & 0.99 & 1.36 & 0.64 & 1.06 & 1.17 \\
\cellcolor{OursRow}\textbf{Ours, one checkpoint} & \cellcolor{OursRow}\textbf{4.62} & \cellcolor{OursRow}0.97 & \cellcolor{OursRow}\textbf{1.79} & \cellcolor{OursRow}\textbf{3.76} & \cellcolor{OursRow}1.79 & \cellcolor{OursRow}\textbf{2.59}
\\
\bottomrule
\end{tabular}

\vspace{5pt}
\setlength{\tabcolsep}{2.4pt}
\begin{tabular}{@{}lrrrrr@{}}
\toprule
\multicolumn{6}{@{}l}{\textbf{Simultaneously activated controls}}\\
Active & All signs$\uparrow$ & Min. effect$\uparrow$ & Add. error$\downarrow$ & Drift$\downarrow$ & DINO$\uparrow$\\
\midrule
Pairs   & 96.7 & 0.356 & 0.065 & 0.349 & 0.878 \\
Triples & 86.1 & 0.240 & 0.116 & 0.406 & 0.822 \\
\bottomrule
\end{tabular}
\end{table}

\begin{figure}[H]
\centering
\includegraphics[width=0.62\linewidth]{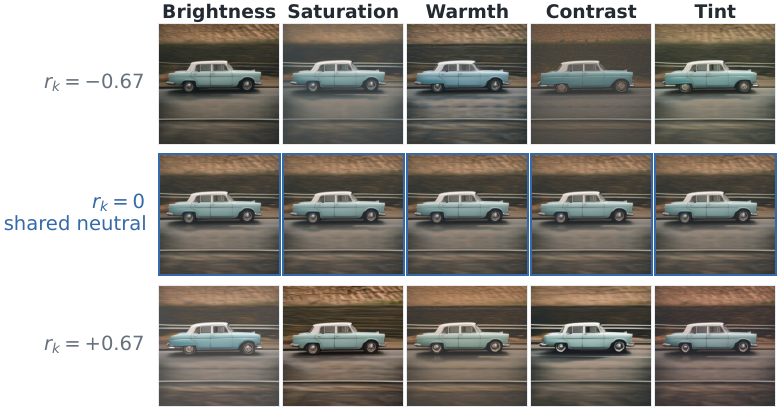}
\caption{\textbf{Five continuous controls from one checkpoint.} Calibrated
branches share an unseen prompt, seed, noise, and sampler.  Rows use
$r_k\in\{-0.67,0,+0.67\}$.  Blue denotes the pixel-identical neutral.}
\label{fig:qualitative}
\end{figure}

\paragraph{One checkpoint exposes individually controllable and composable axes.}
The five-branch checkpoint reaches average selectivity
$\AttributeOursAverage$, compared with $\AttributeBestBaselineAverage$ for the
strongest baseline average, and leads on brightness, contrast, and warmth in the
individual-control panel of Table~\ref{tab:multicontrol}.  Every branch has mean
effect at least $\AttributeMinEffect$, mean $\rho\geq\AttributeMinRho$, and DINO
consistency of $\AttributeOursDino$.  Without joint-activation training, all
requested directions are simultaneously correct in $96.7\%$ of pair outputs,
with a 95\% CI of $[93.8,99.4]$, and $86.1\%$ of triple outputs, with a 95\% CI
of $[80.3,92.0]$.  Pair and triple minimum effects remain $0.356$ and $0.240$
natural ranges, with additivity errors of $0.065$ and $0.116$, respectively, as
reported in the simultaneous-control panel of Table~\ref{tab:multicontrol}.
Figure~\ref{fig:qualitative} shows that these
measured responses correspond to coherent visual traversals from a shared
neutral generation.

\begin{figure}[H]
\centering
\includegraphics[width=0.81\linewidth]{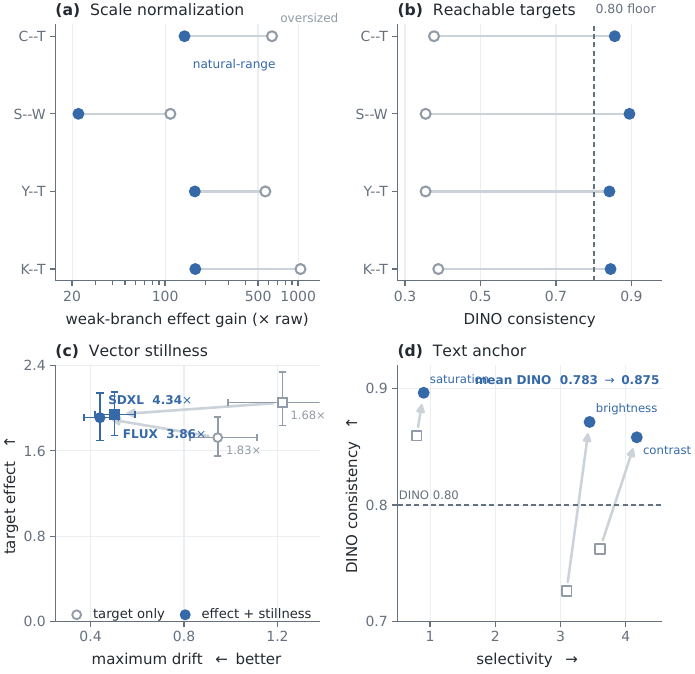}
\caption{\textbf{Component analysis.} Panels a through d evaluate
natural-variation normalization, reachable targets, explicit stillness, and the
frozen-text anchor.  They test weak-coordinate balance, content consistency,
non-target drift, and visual consistency, respectively.}
\label{fig:ablation}
\end{figure}

\paragraph{Component analysis.}
Natural-variation normalization prevents weak-coordinate collapse: across four
measurement pairs, it raises weak-branch response by factors ranging from $22$
to $169$ under
reachable targets, while changing oversized to natural-range targets restores
DINO similarity from the range $0.35$ to $0.39$ to the range $0.84$ to $0.90$.
The zero targets in
Eq.~\eqref{eq:lmetric} reduce maximum drift by $53.4\%$ on SDXL and $58.8\%$ on
FLUX at comparable effect, increasing selectivity from $1.83$ to $4.34$ and
$1.68$ to $3.86$, respectively.  Removing the optional semantic anchor lowers
mean DINO similarity from $0.875$ to $0.783$, as shown in Fig.~\ref{fig:ablation}.
Full
objective averages across brightness, contrast, and saturation are $1.493$
effect, $0.664$ drift, $2.839$ selectivity, and $0.875$ DINO.  Without the text
anchor they become $1.701$, $0.780$, $2.497$, and $0.783$, respectively.

\paragraph{Composition under the full criterion.}
The strict criterion requires correct signs, effects of at least $0.2$ natural
ranges, target errors at most $0.25$, inactive drift at most $0.5$, and DINO of
at least $0.8$.  Under it, $45.2\%$ of pairs and $15.1\%$ of triples pass.  The
gap from directional success isolates failures in magnitude, interference, or
consistency.

\FloatBarrier
\section{Conclusion}
\label{sec:conclusion}

Measured Sliders grounds continuous generative controls in differentiable image
measurements shared by observability, learning, and calibration.  Across SDXL
and FLUX.1-dev, it yields an ordered lighting axis and a five-branch checkpoint
for calibrated appearance control and zero-shot composition.  The pre-training
test predicts all six learnability outcomes, while calibration reduces endpoint
dispersion by nearly $29\times$.  Decoded units separate control quality from
LoRA scale and reveal higher-order branch interference.  Content-adaptive
calibration and broader measurement panels remain open directions.

\clearpage
\begingroup
\let\small\scriptsize
\bibliographystyle{splncs04}
\bibliography{refs}
\endgroup

\end{document}